\documentclass[11pt,a4paper]{article}
\usepackage[hyperref]{naaclhlt2018}
\usepackage{amssymb}
\usepackage{amsmath}
\usepackage{times}
\usepackage{latexsym}
\usepackage{url}
\usepackage[utf8]{inputenc} 
\usepackage{graphicx}

\aclfinalcopy 

\title{BUT-FIT at SemEval-2019 Task 7:  Determining the Rumour Stance with Pre-Trained Deep Bidirectional Transformers}

\author{Martin Fajcik, Lukáš Burget, Pavel Smrz\\
  Brno University of Technology, Faculty of Information Technology\\
  612\,66 Brno, Czech Republic \\
  {\tt \{ifajcik,burget,smrz\}@fit.vutbr.cz} }

\date{}

\begin{document}
\maketitle
\begin{abstract}
This paper describes our system submitted to SemEval 2019 Task 7: RumourEval 2019: Determining Rumour Veracity and Support for Rumours, Subtask A \cite{rumoureval2019}. The challenge focused on classifying whether posts from Twitter and Reddit \emph{support, deny, query,} or \emph{comment} a hidden rumour, truthfulness of which is the topic of an underlying discussion thread. We formulate the problem as a stance classification, determining the rumour stance of a post with respect to the previous thread post and the source thread post. The recent BERT architecture was employed to build an end-to-end system which has reached the F1 score of 61.67\,\% on the provided test data. It finished at the 2\textsuperscript{nd} place in the competition, without any hand-crafted features, only 0.2\,\% behind the winner.
\end{abstract}

\section{Introduction}
Fighting false rumours at the internet is a tedious task. Sometimes, even understanding what an actual rumour is about may prove challenging. And only then one can actually judge its veracity with an appropriate evidence. The works of \cite{ferreira2016emergent, enayet2017niletmrg} focused on prediction of rumour veracity in thread discussions. These works indicated that the veracity is correlated with stances of the discussion participants towards the rumour.
Following this assumption, the participants of the SubTask A in the SemEval competition  Task~7 were asked to classify whether the stance of each post in a given Twitter or Reddit thread \textit{supports}, \textit{denies}, \textit{queries} or \textit{comments} hidden rumour.
Potential applications of such a function are wide, ranging from an analysis of popular events (political discussions, academy awards, etc.) to quickly disproving fake news during disasters. 

\begin{figure}[h!]
\includegraphics[width=0.47\textwidth, angle=0]{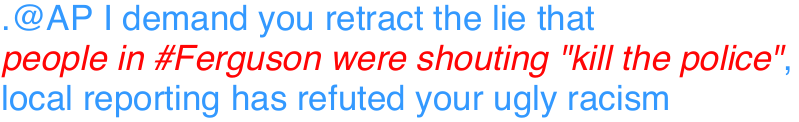}
\vspace{-2ex}
\caption{\label{fig:source_example} An example of discussion's source post denying the actual rumour which is present in the source post -- annotated with the red cursive}
\vspace{-1ex}
\end{figure}

Stance classification (SC) in its traditional form is concerned with determining the attitude of a source text towards a target text \cite{mohammad2016semeval} and it has been studied thoroughly for discussion threads  \cite{walker2012stance, hasan2013stance,chuang2015stance}. 
However, the objective of SemEval 2019 Task 7 is to determine the stance to hidden rumour which is not explicitly given (it can be often inferred from the source post of the discussion -- the root of the tree-shaped discussion thread -- as demonstrated in Figure~\ref{fig:source_example}). The competitors were asked to classify the stance of the source post itself too.

The approach followed in our work builds on recent advances in language representation models. We fine-tune the pre-trained end-to-end Bidirectional Encoder Representations from Transformers (BERT) model~\cite{devlin2018bert}, while using the discussion's source post, target's previous post and the target post itself as inputs to determine the rumour stance of the target post. Our implementation is available online.\footnote{\url{www.github.com/MFajcik/RumourEval2019}}
\section{Related Work}
\textbf{Previous SemEval competitions}: In recent years, there were two SemEval competitions targeting the stance classification. The first one focused on the setting in which the actual rumour was provided \cite{mohammad2016semeval}. The organizers of SemEval-2016 Task~6 prepared a benchmarking system based on SVM using hand-made features and word embeddings from their previous system for sentiment analysis \cite{MohammadKZ2013}, outperfoming all the challenge participants.

The second was previous RumourEval competition won by a system based on word vectors, handcrafted features\footnote{\label{baseline_features}The features included: a flag indicating whether a tweet is a source tweet of a conversation, the length of the tweet, an indicator of the presence of urls and images, punctuation, cosine distance to source tweet and all other tweets in the conversation, the count of negation and swear words, and an average of word vectors corresponding to the tweet.} and an LSTM \cite{hochreiter1997long} summarizing information of the discussion's branches \cite{kochkina2017turing}. Other submissions were either based on similar handcrafted features \cite{singh2017iitp, wang2017ecnu,enayet2017niletmrg}, features based on sets of words for determining language cues such as Belief or Denial \cite{bahuleyan2017uwaterloo}, post-processing via rule-based heuristics after the feature-based classification \cite{srivastava2017dfki}, Convolutional Neural Networks (CNNs) with rules \cite{lozano2017mama}, or end-to-end CNNs that jointly learnt word embeddings \cite{chen2017ikm}.

\textbf{End-to-End approaches}:
\cite{augenstein2016stance} encodes the target text by means of a bidirectional LSTM (BiLSTM), conditioned on the source text and empirically shows that the conditioning on the source text matters. \cite{du2017stance} proposes target augmented embeddings -- embeddings concatenated with an average of the source text embeddings and applies these to compute an attention based on the weighted sum of the target embeddings that were previously transformed via the BiLSTM. \cite{mohtarami2018automatic} proposes an architecture that encodes the source and the target text via a LSTM and a CNN separately and then uses a memory network together with a similarity matrix to capture the similarity between the source and the target text, and infers a fixed-size vector suitable for the stance prediction.

\section{Dataset}
\begin{table}[!ht]
	\centering
\begin{tabular}{|l|c|c|c|c|c|}
\hline
               & \textbf{S} & \textbf{D} & \textbf{Q} & \textbf{C} & \textbf{Total} \\ \hline
\textbf{train} & 925        & 378        & 395        & 3519       & 5217           \\ \hline
in \%          & 18         & 7          & 8          & 67         &                \\ \hline
\textbf{dev}   & 102        & 82         & 120        & 1181       & 1485           \\ \hline
in \%          & 7          & 6          & 8          & 80         &                \\ \hline
\textbf{test}  & 157        & 101        & 93         & 1476       & 1827           \\ \hline
in \%          & 9          & 6          & 5          & 81         &                \\ \hline
\end{tabular}
\caption{\label{tab:dataset_dist}
		Histogram and distribution of examples through classes in the train/dev/test dataset splits. The individual examples belong into 327/38/81 train/dev/test tree-structured discussions.}
\end{table}
Provided dataset was collected from Twitter and Reddit tree-shaped discussions. The stance labels were obtained via crowdsourcing. The Twitter discussions are based on recent popular topics -- Sydney siege, Germanwings crash etc. and there are 9 total topics covered in the training data. The Twitter part of test data contains different topics. The Reddit discussions cover various topics and the discussions are in most cases not related to each other. We provide a deeper insight at dataset in Appendix \ref{app:dataset}.

\section{BUT-FIT's System Description}
\subsection{Preprocessing}
We replace URLs and mentions with special tokens $\$URL\$$ and $\$mention\$$ using tweet-processor\footnote{\url{https://github.com/s/preprocessor}}. We use spaCy\footnote{\url{https://spacy.io/}} to split each post into sentences and add $[EOS]$ token to terminate each sentence. Then we use tokenizer that comes with Hugging Face pytorch re-implementation of BERT\footnote{\label{reimplementation_BERT}\url{https://github.com/huggingface/pytorch-pretrained-BERT}}. The tokenizer lowercases the input and applies the WordPiece encoding \cite{wu2016google} to split input words into most frequent n-grams present in the pre-training corpus, effectively representing text at the sub-word level while keeping only 30,000 token vocabulary.

\subsection{Model}
\begin{figure*}[t!]
	      	\includegraphics[width=1.\textwidth, angle=0]{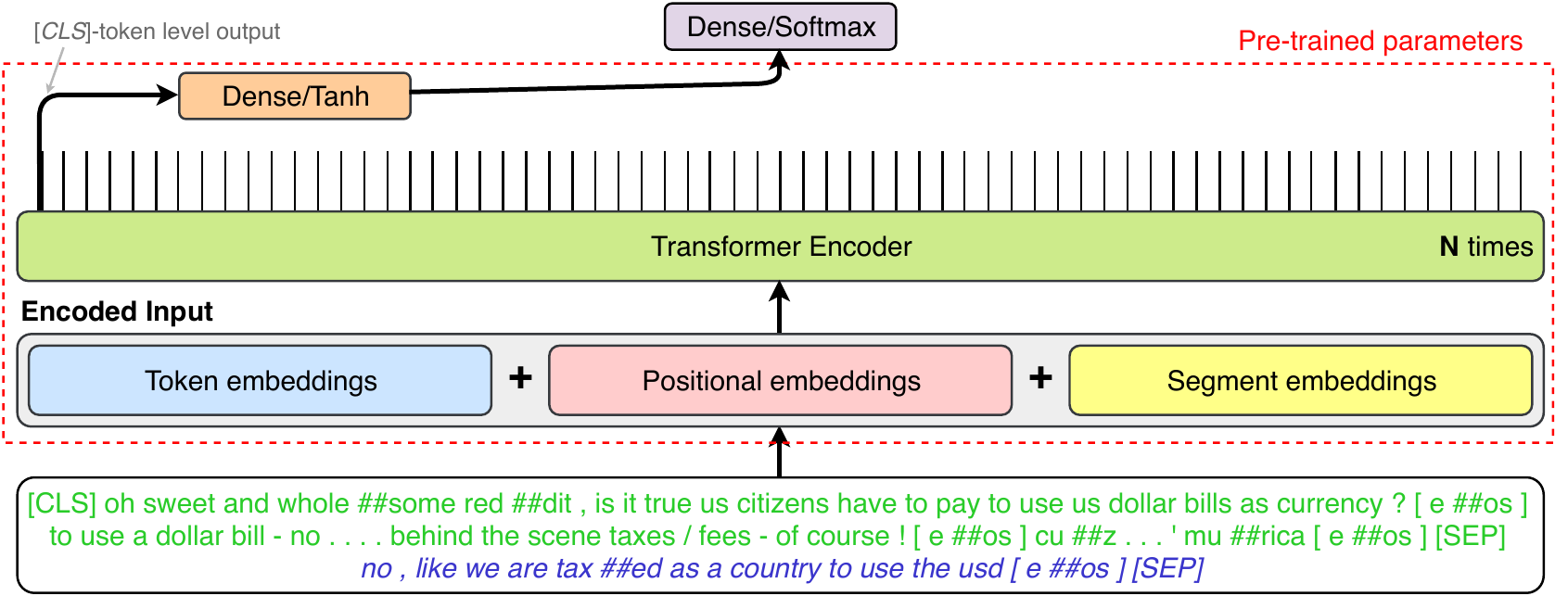}
	      	\caption{\label{fig:architecture} An architecture of BUT-FIT's system. Text segment containing $document_1$ is annotated with green color, segment that contains $document_2$ (target post) is annotated with blue cursive. The input representation is obtained by summing the input embedding matrices $E = E_t + E_s + E_p \in \mathbb{R}^{L \times d}$, with $L$ being the input length and $d$ input dimensionality.  The input is passed $N$ times via transformer encoder.
Finally, the $[CLS]$-token level output is fed via two dense layers yielding the class predictions.}
\vspace{-3ex}
\end{figure*}

Following the recent trend in transfer learning from language models (LM), we employ the pre-trained BERT model. The model is first trained on the concatenation of BooksCorpus  (800M  words) \cite{zhu2015aligning} and  English  Wikipedia  (2,500M  words) using the multi-task objective consisting of LM and machine comprehension (MC) sub-objectives. The LM objective aims at predicting the identity of 15\% randomly masked tokens present at the input\footnote{The explanation of token masking is simplified and we refer readers to read details in the original paper \cite{devlin2018bert}.}. Given two sentences from the corpus, the MC objective is to classify whether the second sentence follows the first sentence in the corpus. The sentence is replaced randomly in half of the cases.
During the pre-training, the input consists of two documents, each represented as a sequence of tokens divided by special $[SEP]$ token and preceeded by $[CLS]$ token used by the MC objective, i.e. $[CLS]document_1[SEP]document_2[SEP]$. The input tokens are represented via jointly learned token embeddings ${E_t}$, segment embeddings $E_s$ capturing whether the word belongs into $document_1$ or $document_2$ and positional embeddings ${E_p}$ since self-attention is position-invariant operation (see \cite{devlin2018bert} for details).

In our solution, we follow the assumption that the stance of the discussion's post depends only on itself, the source thread post and  previous thread post. Since the original input is composed of two documents, we experimented with various ways of encoding the input (Section \ref{input_experiments}) ending up with just a concatenation of source and previous post as $document_1$ (left empty in case of source post being the target post)  and target post as $document_2$.  The discriminative fine-tuning of BERT is done using the $[CLS]$-token level output and passing it via two dense layers yielding the posterior probabilities as depicted in Figure \ref{fig:architecture}. Weighted cross-entropy loss is used to ensure the flat prior over the classes.

\subsection{Ensembling}
Before submission we trained 100 models, which differed just by learning rate. We experimented with 4 different system fusions in order to increase F1 measure and compensate for overfitting:\\
\texttt{TOP-N} fusion chose 1 model randomly to add into the ensemble, then randomly shuffled the rest and tried adding them into ensemble one at the time, while iteratively calculating ensemble's F1 by averaging the output probabilities to approximate the bayesian model averaging. If adding model into ensemble increased the F1, model has been permanently included in the ensemble. The process has been repeated until no further model improving the ensemble's F1 has been found. This resulted into set of 17 best models.\\
\texttt{EXC-N} fusion chose all models into the ensemble and then iterativly dropped one model at the time s.t. dropping it resulted in the largest increase of the ensemble's F1, stopping when dropping any ensemble's model did not increased the F1. Using this approach, we ended up using 94 models.\\
\texttt{TOP-N}$_{s}$ is analogous to \texttt{TOP-N} fusion, but we average the pre-softmax scores instead of output class probabilities.\\
\texttt{OPT-F1} fusion aims at learning weights summing up to 1 for weighted average of the output probabilities from models selected via the procedure used in \texttt{TOP-N}. The weights are estimated using modified Powell's method from SciPy to maximize the F1 score on dev data.

\begin{table*}[ht!]
\begin{center}
\begin{tabular}{|l|c|c|c|c|c|c|c|c|}
\hline
                                              & \#$\Theta$                 & Acc$_{test}$ & macro F1$_{dev}$ & macro F1${_{test}}$                    & F1${_{S}}$ & F1${_{Q}}$ & F1${_{D}}$ & F1${_{C}}$ \\ \hline
Branch-LSTM                                   & 453K                       & $84.10$                        & -          & 49.30                                        & 43.80      & 55.00      &  7.10      & 91.30      \\ \hline
FeaturesNN                                    & 205K                       & $82.84$      & $45.46\pm1\mathrm{e}{-2}$     & $44.55\pm2\mathrm{e}{-2}$ & 40.29      & 40.12      & 17.69      & 80.43      \\ \hline
BiLSTM+SelfAtt                                & 28M                        & $83.59$      & $47.55\pm6\mathrm{e}{-3}$   & $46.81\pm6\mathrm{e}{-3}$   & 42.21      & 45.20      & 17.75      & 81.92      \\ \hline
BERT$_{base}$                                 & 109M                       & $84.67$      & $51.40\pm1\mathrm{e}{-2}$      & $53.39\pm3\mathrm{e}{-2}$& 43.49      & 59.88      & 18.42      & 90.36      \\ \hline
BERT$_{big-noprev}$                           & 335M                       & $84.33$      & $52.61\pm2\mathrm{e}{-2}$      & $52.91\pm4\mathrm{e}{-2}$& 42.37      & 55.17      & 24.44      & 90.15      \\ \hline
BERT$_{big-nosrc}$                            & 335M                       & $84.51$      & $53.72\pm2\mathrm{e}{-2}$      & $55.13\pm3\mathrm{e}{-3}$& 43.02      & 56.93      & 26.53      & 90.51      \\ \hline
BERT$_{big}$                                  & 335M                       & $84.08$      & $56.24\pm9\mathrm{e}{-3}$  & $56.70\pm3\mathrm{e}{-2}$    & 44.29      & 57.07      & 35.02      & 90.41      \\ \hline
BERT$_{big}$ \texttt{EXC-N}$^{*}$             &  -                         & $85.50$      & $58.63$    & $60.28$                         & 48.89      & 62.80      & 37.50      & 91.94      \\ \hline
BERT$_{big}$ \texttt{TOP-N}$^{*}$             &  -                         & $85.22$      & $62.58$    & $60.67$                         & 48.25      & 62.86      & 39.74      & 91.83      \\ \hline
BERT$_{big}$ \texttt{OPT-F1}                  &  -                         & $85.39$      & $62.68$    & $61.27$                         & 48.03      & 62.26      & 42.77      & 92.01      \\ \hline
BERT$_{big}$ \texttt{TOP-N}$_{s}$             &  -                         & $85.50$      & $61.73$    & \textbf{61.67}                  & 49.11      & 64.45      & 41.29      & 91.84      \\ \hline
\end{tabular}
\end{center}
\vspace{-1ex}

\caption{\label{tab:results} Our achieved results. Results for single model were obtained by training at least 10 models and we report mean and standard deviation for these. \#$\Theta$ denotes the number of parameters. The columns F1${_{S}}$ through F1${_{C}}$ contain individual F1 scores for problem classes. All ensemble models are optimized for F1-score on dev data. \texttt{BiLSTM+SelfAtt} contains 4.2M parameters without pre-trained BERT embeddings. BERT$_{big-nosrc}$ and BERT$_{big-noprev}$ denote ablations with empty source or target post respectively. Note that the accuracy is biased towards different training data prior as shown in Table \ref{tab:dataset_dist}. Our SemEval submissions are denoted with $^{*}$. Winning BLCU-nlp system achieved 61.87 F1 score on test data. More available at {\small \url{http://tinyurl.com/y3m5mskd}}.}
\vspace{-3ex}
\end{table*}

\vspace{-1ex}
\section{Experimental Setup}
We implemented our models in pytorch, where we use Hugging Face re-implementation (Footnote \ref{reimplementation_BERT}) in \textit{"bert-large-uncased"} setting pre-trained with $24$ transformer layers, hidden unit size of $d = 1024$, $16$ attention heads and $~335M$ parameters. When building an ensemble, we picked the learning rates from the interval $[1\mathrm{e}{-6},2\mathrm{e}{-6}]$. Each epoch, we iterate over dataset in an ordered manner, starting with shortest sequence as we found this to be helpful. We truncate sequences at maximum length $l=200$ with a heuristic -- firstly we truncate the $document_1$ to length $l/2$, if that is not enough, then we truncate the $document_2$ to the same size. We kept batch size at $32$ and keep other hyperparameters the same as in BERT paper. We use the same Adam optimizer with L2 weight decay of $0.01$ and no warmup. We trained the model on GeForce RTX 2080 Ti GPU.

\section{Results and Analysis}
\label{input_experiments}
We compare our solution with three baselines. The first is \texttt{branch-LSTM} baseline provided by the task organizers\footnote{\url{http://tinyurl.com/y4p5ygn7}} -- inspired by the winning system of RumourEval 2017. The second baseline (\texttt{FeaturesNN}) is our re-implementation of first baseline in pytorch without LSTM -- posts are classified via 2 layer network (ReLU/Softmax) only by features named in  Footnote \ref{baseline_features}. In the third case (\texttt{BiLSTM+SelfAtt}), we used the same input representation as our submitted model, but replaced BERT with 1-layer BiLSTM followed by self-attention and a softmax layer as proposed by \cite{lin2017structured}, except the orthogonality constraint is not used as we did not found it helpful.

The results are shown in Table \ref{tab:results}. Our BERT models encountered high variance of the results during the training. We assume the cause of this might be the problem difficulty, small training set and the model complexity. To counteract, we decided to discard all the models with less than 55 F1 score on dev data and we averaged the output class probability distributions when ensembling. Our initial experiments used sequences up to length 512, but we found no difference when truncating them down to 200.

\textbf{What features weren't helpful}: We tried adding a number of features to the pooled output (after dense/tanh layer) including positive, neutral and negative sentiment and all the features used by \texttt{FeaturesNN} baseline. We also tried adding jointly learned POS, NER and dependency tag embeddings as well as third segment embedding\footnote{We tried adding the learned representations to the input the same way the segment/positional embeddings are added.} or explicit $[SEP]$ token to separate source and previous post in BERT's input without observing any improvement.


\section{Conclusion}
Our approach achieves 61.67 macro F1 score improving over baseline by 12.37\%, while using only discussion's source post, previous post and the target post to classify the target post's stance to rumour. In our case study, we noticed that few examples are not answerable by human while using only these information sources. Therefore, in future we would like to extend our system with relevance scoring system, scoring the all discussion's posts and picking up the most relevant ones to preserve the context of understanding.

\section*{Acknowledgments}
This work was supported by  {\small[Acknowledgments will be filled upon acceptance.]}

\bibliography{semeval2018}
\bibliographystyle{acl_natbib}

\appendix

\section{Supplemental Material}
\label{sec:supplemental}
\subsection{Dataset Insights}
\label{app:dataset}
 For each discussion from Twitter and Reddit, the dataset contains its whole tree structure and metadata, which are different for both sites (e.g. upvotes in Reddit). When analyzing the data, we also uncovered a few anomalies: 
12 data points to do not contain any text and according to organizers they were deleted by users at the time of download and been left it in place so as not to break the conversational structure,
the query stance of few examples taken from subreddit DebunkThis\footnote{https://www.reddit.com/r/DebunkThis/} is strictly dependent on domain knowledge and the strict class of some examples is amibgious and they should probably be labelled with multiple classes.
\subsubsection{Domain knowledge dependency}
Examples from subreddit DebunkThis have all the same format "Debunk this: [statement]", e.g.
\textit{"Debunk this: Nicotine isn't really bad for you, and it's the other substances that makes tobacco so harmful."}. All these examples are labelled as queries. 
\subsubsection{Class ambiguity}
For instance source/previous post \textit{"This is crazy! {\#}CapeTown {\#}capestorm {\#}weatherforecast https://t.co/3bcKOKrCJB"} and target post 
\textit{"@RyGuySA Oh my gosh! Is that not a tornado?! Cause wow, It almost looks like one!"}, officialy labelled in the test data as a comment, but we believe it might be a query as well.
\subsection{Additional Introspection}
\begin{figure*}
\centering
	      	\includegraphics[width=0.7\textwidth, angle=0]{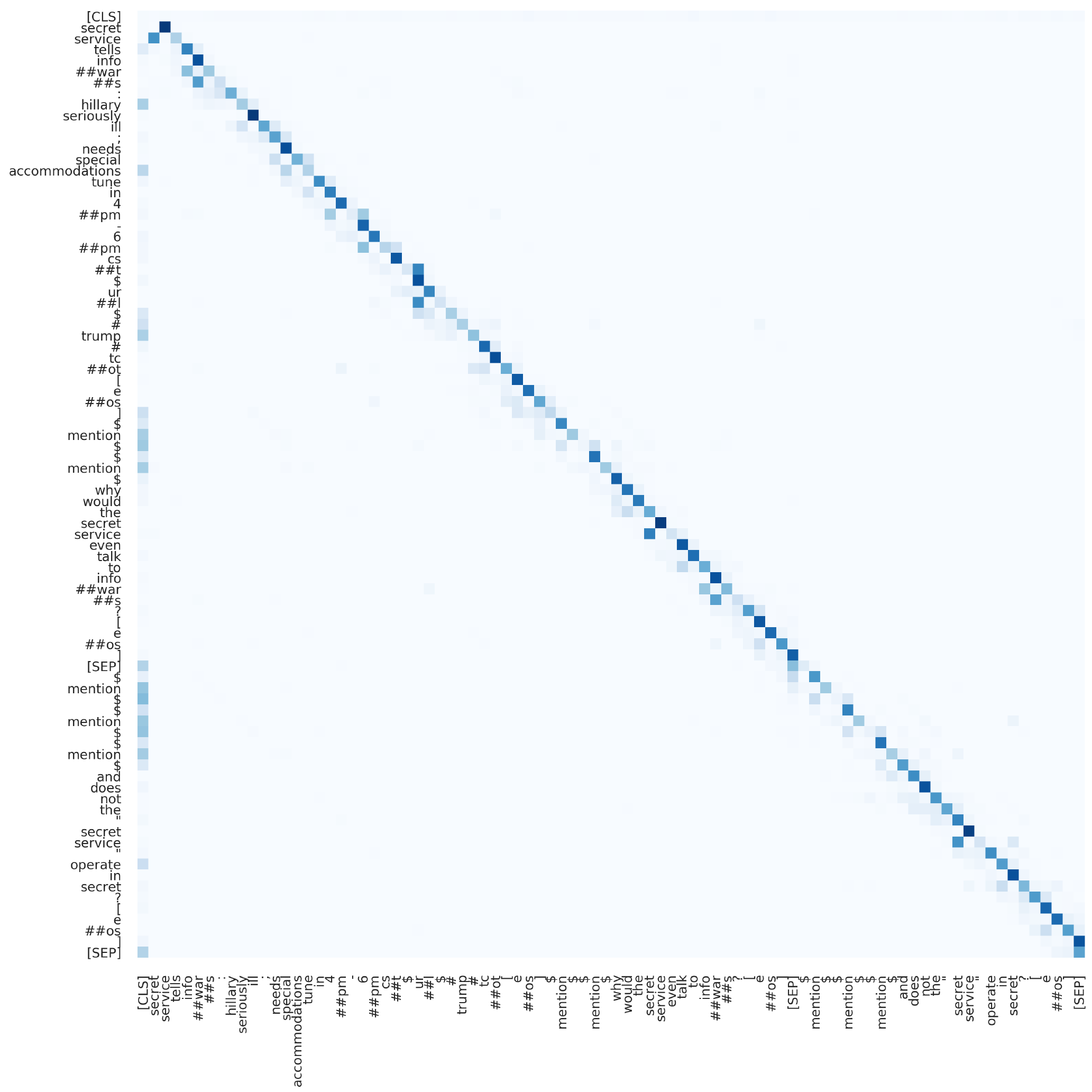}
	      	\caption{\label{fig:att4} 'Hard' local context aggregation -- the signal is mostly sent further to another transformer encoder layer.}
\end{figure*}

The following figures \ref{fig:att1}, \ref{fig:att2}, \ref{fig:att3}, \ref{fig:att4} contain selected insights at the attention matrices $A$ from multi-head attention defined as \eqref{eq:mhatt}, where $Q,K \in \mathbb{R}^{L \times d_k}$ are matrices containing query/value vectors and $d_k$ is the key/value dimension. The insights are selected from the heads at the first layer of transformer encoder.

\begin{equation} \label{eq:mhatt}
A = \frac{QK^\top}{\sqrt{d_k}}
\end{equation}

\begin{figure*}
\centering
	      	\includegraphics[width=0.7\textwidth, angle=0]{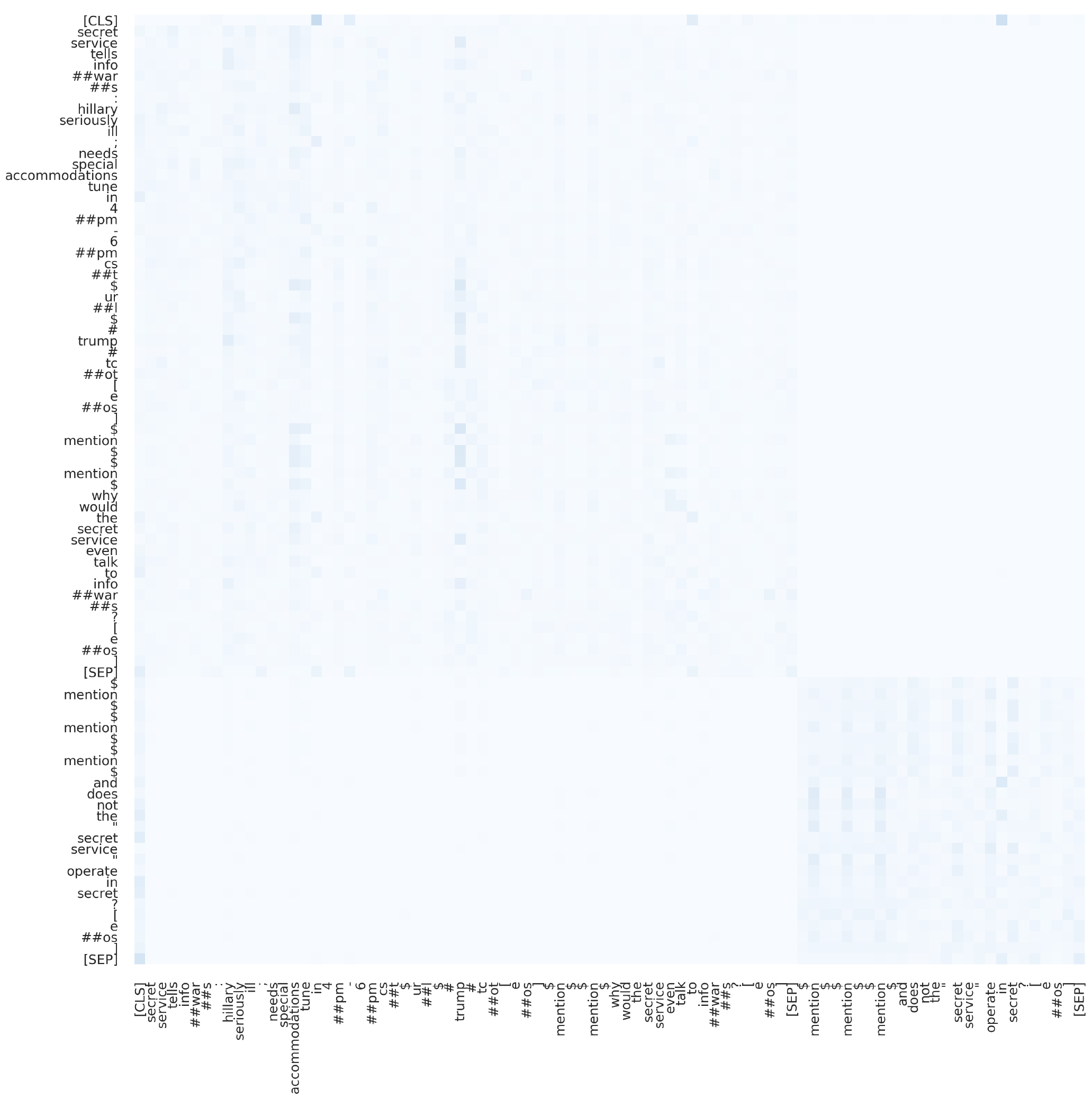}
	      	\caption{\label{fig:att1} Intra-segment attention -- the attention is made only between the subword units from the same segment.}
\end{figure*}

\begin{figure*}
\centering
	      	\includegraphics[width=0.7\textwidth, angle=0]{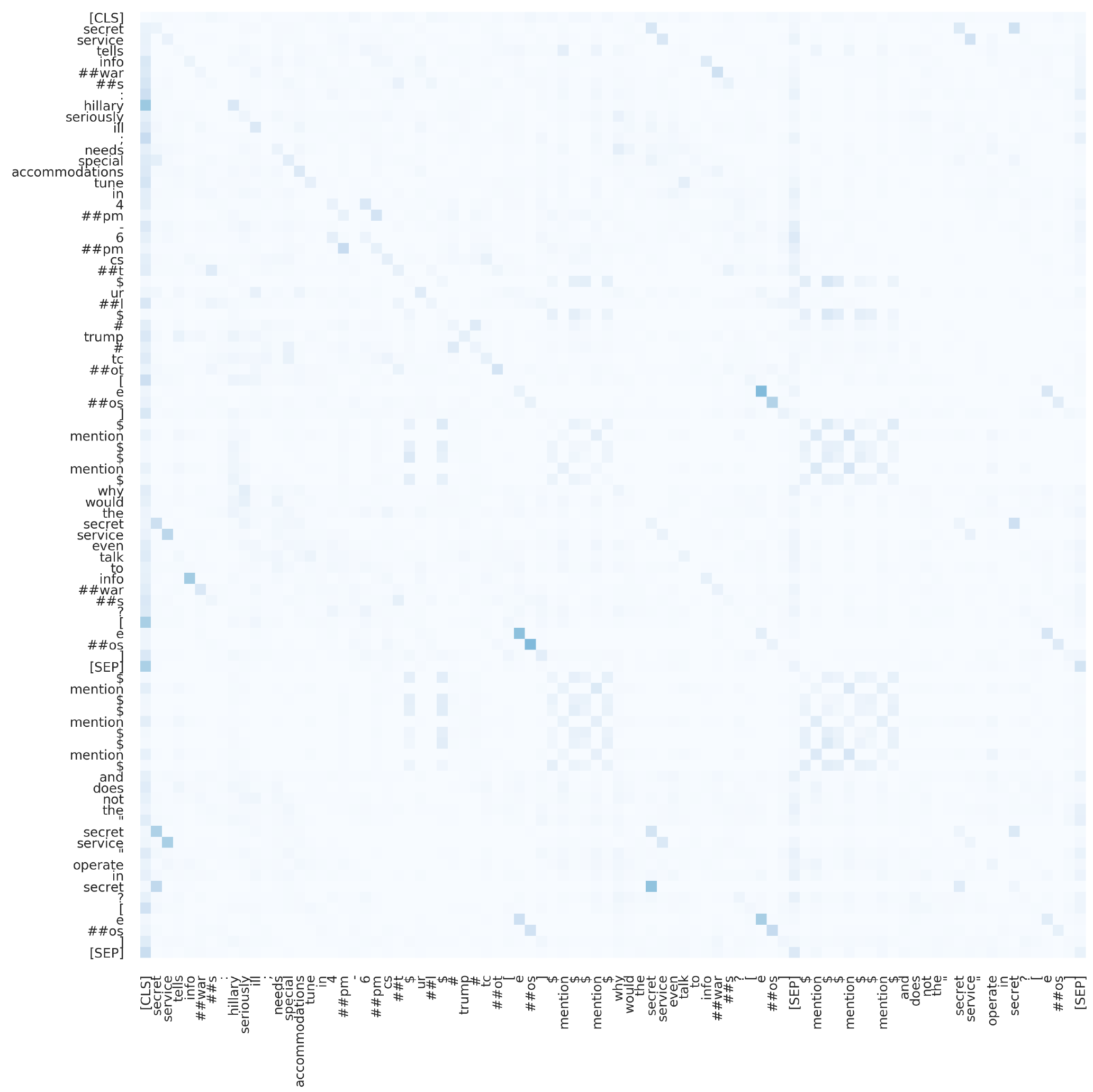}
	      	\caption{\label{fig:att2} Attention matrix capturing the subword similarity.}
\end{figure*}
\begin{figure*}
\centering
	      	\includegraphics[width=0.7\textwidth, angle=0]{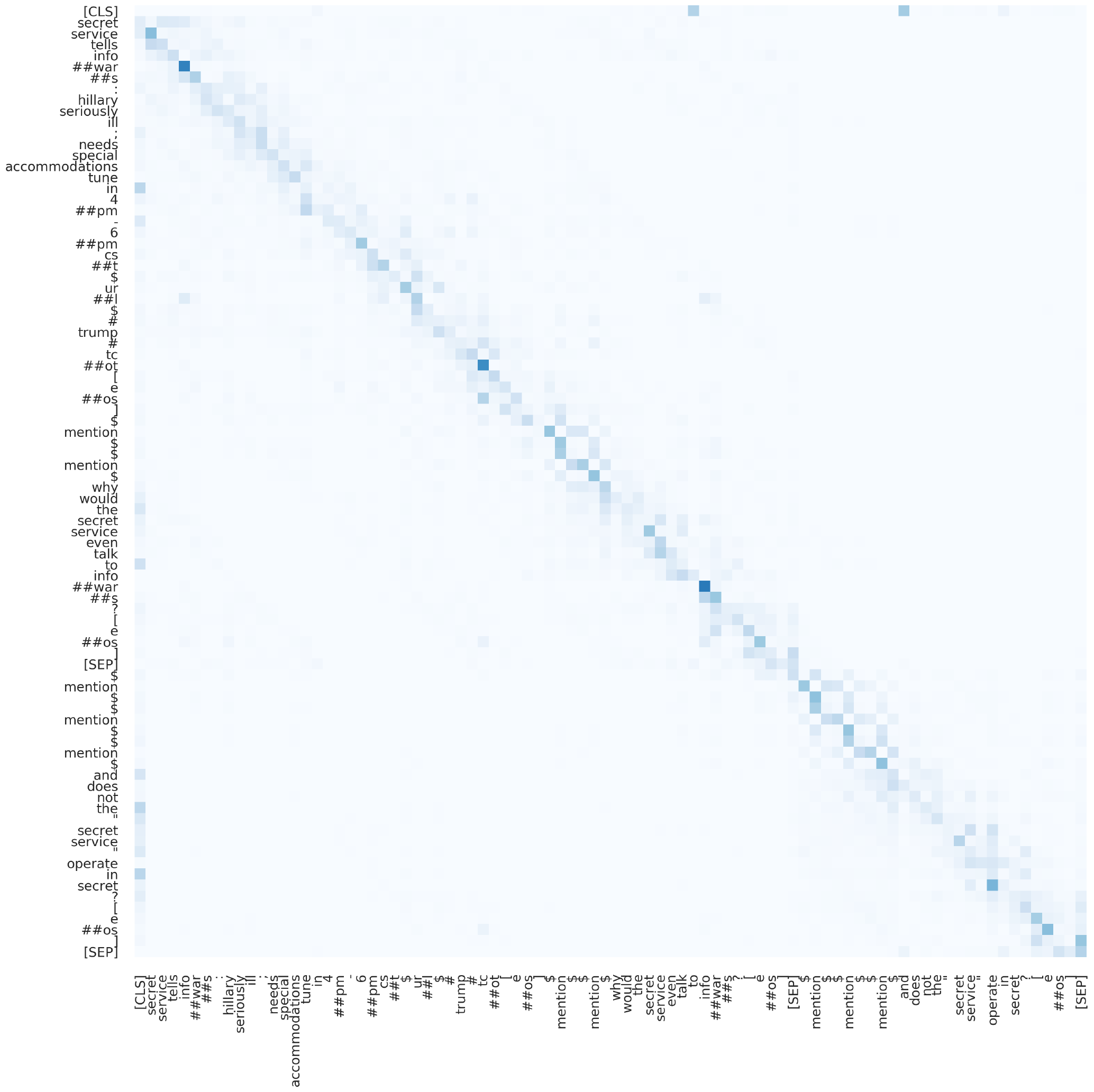}
	      	\caption{\label{fig:att3} 'Soft' local context aggregation.}
\end{figure*}

\end{document}